\documentclass[conference]{IEEEtran}
\IEEEoverridecommandlockouts
\usepackage[utf8]{inputenc}
\usepackage[T1]{fontenc}
\usepackage{amsmath,amssymb}
\usepackage{graphicx}
\usepackage{booktabs}
\usepackage{enumitem}
\usepackage{xcolor}
\usepackage{cite}
\usepackage{url}
\usepackage{microtype}
\usepackage{capt-of}
\usepackage{tikz}
\usetikzlibrary{arrows.meta,positioning,fit}
\usepackage[hidelinks]{hyperref}
\hypersetup{pdftitle={vla.cpp: A Unified Inference Runtime for Vision-Language-Action Models},
pdfauthor={Khanh D. Nguyen, Hung T. Ho, Chinh T. Nguyen, Thanh Q. Duong, Linh D. Le, Duy M. H. Nguyen, Vien A. Ngo, An T. Le}}

\title{\texttt{vla.cpp}: A Unified Inference Runtime for\\Vision-Language-Action Models}
\author{\IEEEauthorblockN{
Khanh D. Nguyen\textsuperscript{1}, Hung T. Ho\textsuperscript{1},
Chinh T. Nguyen\textsuperscript{1}, Thanh Q. Duong\textsuperscript{1},\\
Linh D. Le\textsuperscript{1}, Duy M. H. Nguyen\textsuperscript{4,5,6},
Vien A. Ngo\textsuperscript{1,2}, An T. Le\textsuperscript{1,2,3}}
\thanks{\textsuperscript{1}VinRobotics, Vietnam.
\textsuperscript{2}Center for AI Research, VinUniversity, Vietnam.
\textsuperscript{3}Intelligent Autonomous Systems, TU Darmstadt, Germany.
\textsuperscript{4}Max Planck Research School for Intelligent Systems, Germany.
\textsuperscript{5}University of Stuttgart, Germany.
\textsuperscript{6}German Research Center for Artificial Intelligence.
Corresponding author: An T. Le, \texttt{an@robot-learning.de}.}}

\IEEEaftertitletext{
\definecolor{vcblue}{RGB}{31,101,160}
\definecolor{vcorange}{RGB}{214,118,36}
\definecolor{vcgreen}{RGB}{34,132,94}
\definecolor{vcred}{RGB}{186,58,52}
\definecolor{vcgray}{RGB}{115,115,115}

\newcommand{\pbx}[1]{(#1-76)*0.126}          
\newcommand{\pby}[1]{(#1-76)*0.12}           
\newcommand{\bary}[1]{(#1)*0.0068182}        
\newcommand{\rlx}[1]{(log10(#1)-2)*2.0287}   
\newcommand{\rly}[1]{(#1-30)*0.0428571}      

\newsavebox{\teaserbox}
\sbox{\teaserbox}{%
\begin{tikzpicture}[
    x=1cm, y=1cm,
    ptitle/.style={font=\scriptsize\bfseries, anchor=south west, inner sep=0pt},
    axline/.style={vcgray, line width=0.4pt},
    tik/.style={vcgray, line width=0.4pt},
    tlab/.style={font=\tiny, text=vcgray, inner sep=1pt},
    alab/.style={font=\tiny, inner sep=1pt},
    note/.style={font=\tiny, text=vcgray, inner sep=1pt, align=left},
    fanbox/.style={draw=vcgray!70, rounded corners=1.5pt, fill=black!3,
                   font=\tiny, align=center, inner sep=2.5pt, text width=3.5cm},
]

\begin{scope}[shift={(0,0)}]
  \node[ptitle] at (0,3.9) {(a) Shared runtime and model format};

  \node[draw=vcblue, fill=vcblue!10, rounded corners=2pt, align=center,
        font=\tiny, text width=1.1cm, inner sep=2.5pt] (bin) at (0.64,2.10)
       {\textbf{\texttt{vla.cpp}}\\[1pt] C++ server\\ $+$ GGUF\\ weights};

  \node[fanbox] (mdl) at (3.58,3.24)
       {\textbf{11 VLA models}\\
        SmolVLA\,\textperiodcentered\,Evo-1\,\textperiodcentered\,$\pi_0$\,\textperiodcentered\,$\pi_{0.5}$\\
        GR00T-N1.5/1.6/1.7\,\textperiodcentered\,VLA-JEPA\\
        VLA-Adapter\,\textperiodcentered\,OpenVLA-OFT\,\textperiodcentered\,BitVLA};

  \node[fanbox] (bck) at (3.58,2.21)
       {\textbf{5 backends}\\
        CUDA\,\textperiodcentered\,Metal\,\textperiodcentered\,CPU\,\textperiodcentered\,SYCL\,\textperiodcentered\,OpenVINO};

  \node[fanbox] (dev) at (3.58,1.00)
       {\textbf{Target hardware}\\[1.5pt]
        Apple M4\\[1.5pt]
        RTX 30xx--50xx\,\textperiodcentered\,Intel Arc A380\\[1.5pt]
        Core Ultra iGPU\,\textperiodcentered\,AGX Orin\,\textperiodcentered\,Orin Nano};

  \draw[-{Latex[length=1.3mm]}, vcblue, line width=0.5pt]
        (bin.east) to[out=35,in=180] (mdl.west);
  \draw[-{Latex[length=1.3mm]}, vcblue, line width=0.5pt]
        (bin.east) to[out=0,in=180]  (bck.west);
  \draw[-{Latex[length=1.3mm]}, vcblue, line width=0.5pt]
        (bin.east) to[out=-35,in=180] (dev.west);

  \node[font=\tiny, text=vcgreen, inner sep=1pt] at (2.65,0.15)
       {model adapters share backend operations};
\end{scope}

\begin{scope}[shift={(5.85,0)}]
  \node[ptitle] at (0,3.9) {(b) Task success vs.\ reference};

  \begin{scope}[shift={(0.85,0.65)}]
    \begin{scope}
      \clip (0,0) rectangle (3.15,3.0);
      \fill[vcblue!12] (0,{-4*0.12}) -- (3.15,{(25-4)*0.12})
                    -- (3.15,{(25+4)*0.12}) -- (0,{4*0.12}) -- cycle;
      \draw[vcblue, line width=0.5pt, densely dashed] (0,0) -- (3.15,3.0);
    \end{scope}
    \draw[axline] (0,0) rectangle (3.15,3.0);

    \foreach \v in {80,90,100}{
      \draw[tik] ({\pbx{\v}},0) -- ({\pbx{\v}},-0.07);
      \node[tlab, anchor=north] at ({\pbx{\v}},-0.07) {\v};
      \draw[tik] (0,{\pby{\v}}) -- (-0.07,{\pby{\v}});
      \node[tlab, anchor=east] at (-0.07,{\pby{\v}}) {\v};
    }
    \node[tlab, anchor=north] at (1.575,-0.32) {reference success rate (\%)};
    \node[tlab, anchor=south, rotate=90] at (-0.50,1.5) {\texttt{vla.cpp} success (\%)};

    \foreach \rf/\ou in {96.0/96, 97.7/96, 79.0/82, 98.2/96, 95.0/95,
                         99.0/99, 99.6/100, 99.2/96, 98.4/97, 99.6/100}{
      \fill[vcblue] ({\pbx{\rf}},{\pby{\ou}}) circle (1.5pt);
    }
    \fill[vcred] ({\pbx{99.0}},{\pby{88}}) circle (1.9pt);
    \node[alab, text=vcred, anchor=north east, align=right]
         at ({\pbx{99.0}-0.06},{\pby{88}+0.05}) {GR00T-N1.6\\(88\% vs.\ 99\%)};

    \node[alab, anchor=west] at ({\pbx{79.0}+0.09},{\pby{82}}) {$\pi_0$};
    \node[alab, anchor=west, align=left] (hundred) at (0.82,2.74)
         {BitVLA, VLA-JEPA\\$\mathbf{100\%}$ SR};
    \draw[vcgray, line width=0.35pt]
         (hundred.east) -- ({\pbx{99.6}-0.05},{\pby{100}+0.02});
    \node[tlab, anchor=south east, align=right, text=vcblue] at (3.07,0.10)
         {$\pm4$\,pp band:\\all but one inside};
  \end{scope}
\end{scope}

\begin{scope}[shift={(10.20,0)}]
  \node[ptitle] at (0,3.9) {(c) Ternary execution};

  \begin{scope}[shift={(0.70,0.65)}]
    \draw[axline] (0,0) -- (2.6,0);
    \draw[axline] (0,0) -- (0,3.0);
    \foreach \v in {0,100,200,300,400}{
      \draw[tik] (0,{\bary{\v}}) -- (-0.06,{\bary{\v}});
      \node[tlab, anchor=east] at (-0.06,{\bary{\v}}) {\v};
    }
    \node[tlab, anchor=south, rotate=90] at (-0.48,1.5) {ms / executed action};

    \fill[vcorange!75] (0.30,0) rectangle (0.64,{\bary{172.8}});
    \fill[vcorange!75] (1.58,0) rectangle (1.92,{\bary{406.6}});
    \fill[vcblue]      (0.68,0) rectangle (1.02,{\bary{37.85}});
    \fill[vcblue]      (1.96,0) rectangle (2.30,{\bary{101.11}});

    \node[tlab, anchor=south] at (0.47,{\bary{172.8}}) {172.8};
    \node[tlab, anchor=south] at (0.85,{\bary{37.85}}) {37.9};
    \node[tlab, anchor=south] at (1.75,{\bary{406.6}}) {406.6};
    \node[tlab, anchor=south] at (2.13,{\bary{101.11}}) {101.1};

    \node[alab, text=vcgreen, anchor=south] at (0.85,{\bary{37.85}+0.20})
         {$\mathbf{4.6\times}$};
    \node[alab, text=vcgreen, anchor=south] at (2.13,{\bary{101.11}+0.20})
         {$\mathbf{4.0\times}$};

    \node[tlab, anchor=north] at (0.66,-0.06) {RTX 3060};
    \node[tlab, anchor=north] at (1.94,-0.06) {AGX Orin};

    \fill[vcorange!75] (0.06,2.80) rectangle (0.24,2.92);
    \node[tlab, anchor=west] at (0.28,2.86) {\texttt{dp4a}};
    \fill[vcblue] (0.06,2.58) rectangle (0.24,2.70);
    \node[tlab, anchor=west] at (0.28,2.64) {IMMA (ours)};
  \end{scope}
  \node[note, anchor=north west] at (0.18,0.13)
       {$S=8$; unchanged task success};
\end{scope}

\begin{scope}[shift={(13.72,0)}]
  \node[ptitle] at (0,3.9) {(d) On the robot: latency vs.\ success};

  \begin{scope}[shift={(0.85,0.65)}]
    \draw[axline] (0,0) rectangle (3.25,3.0);
    \foreach \v in {100,300,1000,3000}{
      \draw[tik] ({\rlx{\v}},0) -- ({\rlx{\v}},-0.07);
      \node[tlab, anchor=north] at ({\rlx{\v}},-0.07) {\v};
    }
    \foreach \v in {40,60,80,100}{
      \draw[tik] (0,{\rly{\v}}) -- (-0.07,{\rly{\v}});
      \node[tlab, anchor=east] at (-0.07,{\rly{\v}}) {\v};
    }
    \node[tlab, anchor=north] at (1.625,-0.32) {round-trip latency (ms, log)};
    \node[tlab, anchor=south, rotate=90] at (-0.50,1.5) {task success (\%)};

    \foreach \ms/\sr/\lo/\hi in {111.4/75/53.1/88.8, 582.0/60/38.7/78.1,
                                 2164.4/60/38.7/78.1, 3652.9/70/48.1/85.5}{
      \draw[vcblue!55, line width=0.6pt]
           ({\rlx{\ms}},{\rly{\lo}}) -- ({\rlx{\ms}},{\rly{\hi}});
      \draw[vcblue!55, line width=0.6pt]
           ({\rlx{\ms}-0.06},{\rly{\lo}}) -- ({\rlx{\ms}+0.06},{\rly{\lo}});
      \draw[vcblue!55, line width=0.6pt]
           ({\rlx{\ms}-0.06},{\rly{\hi}}) -- ({\rlx{\ms}+0.06},{\rly{\hi}});
      \fill[vcblue] ({\rlx{\ms}},{\rly{\sr}}) circle (1.9pt);
    }

    \node[alab, anchor=south west] at ({\rlx{111.4}+0.06},{\rly{75}+0.02}) {CUDA};
    \node[alab, anchor=south west] at ({\rlx{582.0}+0.06},{\rly{60}+0.02}) {Metal};
    \node[alab, anchor=north east] at ({\rlx{2164.4}-0.07},{\rly{60}-0.02}) {CPU i9};
    \node[alab, anchor=south east] at ({\rlx{3652.9}-0.04},{\rly{70}+0.02}) {CPU Ryzen};

    \node[tlab, anchor=north] at ({\rlx{111.4}+0.16},{\rly{53}-0.04}) {17.7\,Hz};
    \node[tlab, anchor=north] at ({\rlx{3652.9}-0.20},{\rly{48}-0.04}) {5.0\,Hz};

    \draw[{Latex[length=1.1mm]}-{Latex[length=1.1mm]}, vcgray, line width=0.4pt]
         ({\rlx{111.4}},0.24) -- ({\rlx{3652.9}},0.24);
    \node[tlab, anchor=center, fill=white, inner sep=1.5pt] at (1.63,0.24)
         {$33\times$ latency};
  \end{scope}
  \node[note, anchor=north west] at (0.08,0.13)
       {20 episodes/backend; $95\%$ CIs};
\end{scope}

\end{tikzpicture}%
}

\begin{minipage}{\textwidth}
\centering
\usebox{\teaserbox}
\captionof{figure}{\texttt{vla.cpp} across models and hardware.
(a) Shared runtime components support eleven VLA models; backend coverage varies by model.
(b) LIBERO-Object success rate (SR) over 100 episodes per model versus published or reproduced reference rates (Table~\ref{tab:fidelity}); the shaded band marks $\pm4$ percentage points.
(c) Tensor-core execution reduces BitVLA's client inference time per executed action by $4.0$--$4.6\times$ at $S=8$.
(d) UR10e task success and 95\% Wilson intervals across four backends; round-trip latency varies by $33\times$.}
\label{fig:teaser}
\end{minipage}
\par\medskip
}

\begin{document}
\maketitle

\begin{abstract}
Deploying vision--language--action (VLA) models on robots requires adapting model-specific inference pipelines to heterogeneous processors and limited onboard memory.
We present \texttt{vla.cpp}, a unified C++ inference runtime for eleven VLA models, with no PyTorch dependency for model execution.
The runtime shares model loading, tensor execution, and serving while retaining architecture-specific attention, conditioning, and action heads.
Iterative policies reuse observation-dependent computation across solver steps, while regression policies predict actions directly.
We evaluate task success on LIBERO-Object and profile supported configurations on NVIDIA, Apple, and Intel hardware.
BitVLA completes 200/200 LIBERO-Object episodes on an 8\,GB Jetson Orin Nano.
A ternary tensor-core kernel accelerates its client inference by $4.0$--$4.6\times$ over the CUDA-core baseline on RTX 3060 and AGX Orin.
A SmolVLA case study links positional-index precision to gripper commands and task success, showing why fixed-input numerical checks should accompany rollout evaluation.
Deployments on UR10e and ALOHA demonstrate physical robot integration; delay and execution-horizon studies characterize synchronous chunked control.
The results demonstrate a common deployment path across VLA architectures and hardware, with numerical validation and control settings guiding deployment alongside inference efficiency.
\end{abstract}

\section{Introduction}
\label{sec:intro}
Vision--language--action (VLA) policies map camera observations and language instructions to robot actions.
Their pipelines combine preprocessing, neural-network execution, and action conversion, often inside model-specific training software.
On robots, these pipelines share limited memory and compute with sensing and control.
Moving a policy out of its training stack requires efficient execution while preserving its numerical behavior.

Portable inference engines have made language models accessible across CPUs and accelerators~\cite{llamacpp,mlcllm}.
VLAs extend this problem to continuous control.
A vision--language \emph{backbone} encodes observations; an \emph{action head} predicts a command sequence, or \emph{action chunk}.
Iterative heads refine an initially random chunk~\cite{pi0paper,smolvla,groot}; regression heads predict it directly~\cite{openvlaoft,bitvla,vlaadapter}.
The robot executes a prefix of the chunk before observing again.
A common runtime must accommodate these attention patterns, prediction procedures, and control interfaces.

We introduce \texttt{vla.cpp}, a unified C++ inference runtime for eleven VLA models, built on \texttt{ggml} and the \texttt{llama.cpp} ecosystem (Fig.~\ref{fig:teaser}).
Model adapters specify each policy's computation, and hardware backends execute shared tensor operations.
Iterative heads reuse observation features across solver steps, while a client manages sensing and control.
Model execution does not depend on PyTorch.
Fixed-input checks complement task evaluation: our SmolVLA case study links positional-index precision to gripper commands and manipulation success.

Our contributions are:
\begin{enumerate}[leftmargin=*,nosep]
\item A unified runtime that shares execution infrastructure across eleven VLA models while preserving model-specific attention and action generation.
\item A numerical validation procedure and case study linking positional-index precision to predicted actions and task success.
\item A systems evaluation across processors and two robot platforms, including memory-constrained Jetson deployment, ternary tensor-core acceleration, and studies of quantization and chunked control.
\end{enumerate}

\section{Related Work}
\label{sec:related}
\paragraph{Policies and deployment systems}
OpenVLA~\cite{openvla} predicts discrete action tokens; $\pi_0$/$\pi_{0.5}$~\cite{pi0paper,pi05paper}, SmolVLA~\cite{smolvla}, Evo-1~\cite{evo1}, and GR00T~\cite{groot} use continuous action generation.
OpenVLA-OFT~\cite{openvlaoft} and VLA-Adapter~\cite{vlaadapter} study parallel prediction, BitVLA~\cite{bitvla} uses ternary weights, and VLA-JEPA~\cite{vlajepa} adds latent world-model training.
Our runtime deploys trained policies without retraining.
LiteVLA-Edge~\cite{litevla} and Lite VLA~\cite{litevlarpi} explore quantized deployment through \texttt{llama.cpp}; ActQuant~\cite{actquant} combines action-guided quantization with a native C/C++ pipeline for OpenVLA-OFT and $\pi_{0.5}$.
VLAgents~\cite{vlagents} addresses serving, and OxyGen~\cite{oxygen} manages attention caches under multitask inference.
Our focus is shared execution across VLA architectures and processor backends, with numerical checks and robot evaluation.

\paragraph{Inference and hardware}
\texttt{llama.cpp}~\cite{llamacpp} provides portable inference and the GGUF weight format; MLC-LLM~\cite{mlcllm} pursues deployment through compilation.
TensorRT and TensorRT Edge-LLM~\cite{tensorrt,tensorrtedgellm} optimize NVIDIA execution, including Jetson Orin support in Edge-LLM.
Ladder and BitBLAS~\cite{ladder,bitblas} develop low-bit multiplication on tensor cores, NVIDIA's specialized matrix units.
VLA-Perf~\cite{vlaperf} and edge characterizations~\cite{charedge} analyze backbone and action-expert costs.
Our profiles, solver sweep, and ternary kernel comparison measure these costs within one runtime.

\paragraph{Chunked control}
Diffusion Policy~\cite{diffusionpolicy} and ACT~\cite{aloha} use action chunks for temporal consistency.
Real-Time Chunking~\cite{rtc} studies inference delay and execution horizon, including injected delays and dynamic tasks; FASTER~\cite{faster} addresses real-time flow-policy execution.
We evaluate synchronous tabletop control, in which the client executes a selected number of actions from each chunk and then waits for the next prediction, complementing methods that overlap inference and execution.

\section{Runtime Design}
\label{sec:engine}
\begin{figure*}[!t]
\centering
\includegraphics[width=0.88\textwidth]{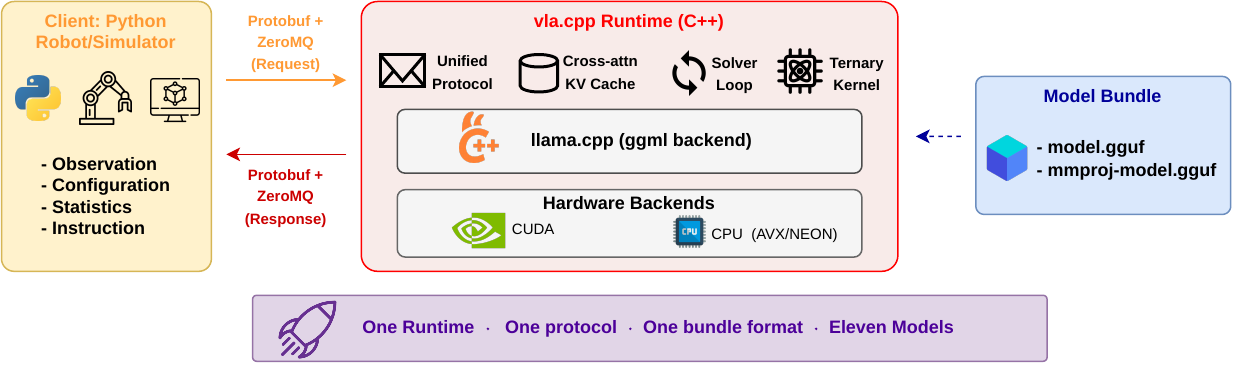}
\caption{Architecture of \texttt{vla.cpp}. The C++ server loads GGUF weights and returns action chunks; the client owns sensing and control. Adapters select model-specific computation. CUDA and CPU illustrate the backend layer; Metal, SYCL, and OpenVINO use the same runtime interface.}
\label{fig:arch}
\end{figure*}
The runtime separates model adapters, tensor-execution backends, and robot clients (Fig.~\ref{fig:arch}).
Adapters define policy-specific graphs; backends execute their operators; clients handle sensing and action replay.

\subsection{Policy interface and action generation}
\label{sec:background}
An observation $o=(I_{1:V},\ell,s)$ contains $V$ camera images, tokenized instruction $\ell$, and robot state $s$.
A policy with parameters $\theta$ predicts an action chunk $A\in\mathbb{R}^{H\times d_a}$ of $H$ actions with $d_a$ command dimensions.
The client executes the first $S$ actions, with $1\leq S\leq H$, before observing again: $H$ is the \emph{prediction horizon}, and $S$ is the \emph{execution horizon}.
An iterative head uses $T\geq1$ \emph{solver steps} to generate one chunk.

Iterative policies compute observation-dependent conditioning $C=F_\theta(o)$ and refine an initial noise sample $X^0\in\mathbb{R}^{H\times d_a}$.
For a flow-matching policy~\cite{flowmatching}, Euler integration gives
\begin{equation}
 X^{k+1}=X^k+(\tau_{k+1}-\tau_k)v_\theta(X^k,\tau_k,C,s),
 \label{eq:euler}
\end{equation}
for $k=0,\ldots,T-1$.
Here $v_\theta$ is the learned velocity field and $\tau_0,\ldots,\tau_T$ is its integration schedule; $\tau_k$ describes generation, not physical execution time.
Schedules and noise follow each checkpoint's conventions: the implemented SmolVLA and $\pi_0$~\cite{openpi} paths integrate $1\to0$, GR00T-N1.7 integrates $0\to1$, and Evo-1 uses uniform initial noise.
Reversing the time coordinate requires the corresponding velocity sign change.
After $T$ steps, output transforms convert $X^T$ to actions.
BitVLA, OpenVLA-OFT, and VLA-Adapter instead regress $A=R_\theta(o)$ directly.
Both paths return a chunk and its dimensions through the same interface.

\subsection{Observation reuse and attention}
\label{sec:components}
Observation tokens form a \emph{prefix}; candidate-action tokens form a \emph{suffix}.
The supported iterative policies keep prefix computation independent of the changing actions, allowing its features and projections to be reused across solver steps.
For $\pi_0$, each layer caches prefix keys and values for joint attention with the changing suffix.
SmolVLA alternates joint- and cross-attention, with extra projections for cross-attention.
GR00T-style heads cache projections of backbone features in a separate action transformer.
These architecture-specific tensors form the collection $C$.

For a conditioning sequence $Z$ in $C$, a cross-attention head caches $K_C=ZW_K$ and $V_C=ZW_V$ and computes
\begin{equation}
 \operatorname{Attn}(Q,K_C,V_C;M)
 =\operatorname{softmax}\!\left(\frac{QK_C^\top}{\sqrt{d_k}}+M\right)V_C.
 \label{eq:attn}
\end{equation}
The equation describes one attention head: $Q$ contains action queries, $W_K,W_V$ are learned projections, and $d_k$ is the key dimension.
The additive mask $M$ is zero for visible keys and $-\infty$ for masked keys.
Softmax operates over keys.
Joint-attention layers append suffix keys and values; adapters preserve model-specific masks, positions, and state-token placement.

Cached conditioning is reused only within a request; each new observation refreshes its values.
For unchanged shapes, graphs and buffers persist across requests, avoiding construction and allocation overhead.
Supported backends also reduce dispatch overhead through compilation or graph capture, which records GPU operations for replay.
If conditioning takes time $P$ and each expert step takes time $E$, model latency is approximately
\begin{equation}
 L_{\mathrm{model}}(T)\approx P+TE.
 \label{eq:cost}
\end{equation}
At fixed shape and precision, the idealized speedup over recomputing conditioning at every solver step is $T(P+E)/(P+TE)$, between $1$ and $T$ for positive costs, excluding transfer and scheduling.

\subsection{Packaging and robot integration}
\label{sec:arch}
GGUF stores weights, architecture metadata, and supported normalization statistics; vision weights may occupy a companion file.
A new policy adds an adapter, converter, and runtime registration.
CPU, CUDA, Metal, SYCL, and OpenVINO builds share tensor interfaces, with model coverage determined by operator support and precision.

The C++ server retains model weights and handles synchronous requests through ZeroMQ and Protocol Buffers.
Requests contain images, token IDs, state, and optional noise and masks; responses contain action chunks, dimensions, and timings, or an error.
A client handles tokenization, sensing, command mapping, and replay.
Input normalization and inverse output transforms follow the reference pipeline, assigned to the adapter or client to avoid duplicate or omitted transforms.
Clients may use Python; model execution is C++.

\subsection{Ternary tensor-core execution}
\label{sec:ternary}
BitVLA weights use scaled ternary values $q_{nj}\in\{-1,0,1\}$, packed four two-bit codes per byte.
The CUDA path quantizes each activation row $x_m$ to signed 8-bit integers:
\begin{equation}
 \begin{aligned}
 \alpha_m&=\frac{\max(\lVert x_m\rVert_\infty,10^{-5})}{127},\\
 a_{mj}&=\operatorname{clip}_{[-128,127]}\!\left(\operatorname{round}\frac{x_{mj}}{\alpha_m}\right),\\
 y_{mn}&=\alpha_m\beta_{g(n)}\sum_j a_{mj}q_{nj}.
 \end{aligned}
 \label{eq:quant}
\end{equation}
Here $m$ indexes activation rows, $j$ and $n$ index input and output channels, and $\beta_{g(n)}$ is the checkpoint scale for output group $g(n)$.
Packed weight tiles are decoded in shared memory; activation tiles are reused across outputs.
Products accumulate in 32-bit integers and are scaled to bfloat16 (BF16), a 16-bit floating-point format.
Rounding uses nearest-even ties; the output equation describes the scaled product before BF16 conversion.
The baseline uses \texttt{dp4a} dot products on CUDA cores; our multi-row path uses integer matrix-multiply--accumulate (IMMA) tensor-core instructions, adapting packed-weight techniques~\cite{ladder,bitblas,bitnetcpp} to BitVLA's shapes and activation conventions.

\subsection{Numerical validation}
\label{sec:validation}
The prediction-check tool fixes weights, images, tokens, state, noise, masks, schedules, and transforms while comparing runtime actions $A$ with reference actions $\widehat A$.
Intermediate tensors localize discrepancies.
We report maximum absolute and relative root-mean-square (RMS) error:
\begin{equation}
 e_\infty=\max_i|A_i-\widehat A_i|,\qquad
 e_{\mathrm{rel}}=\frac{\lVert A-\widehat A\rVert_2}{\lVert\widehat A\rVert_2}.
 \label{eq:error}
\end{equation}
The norms are Euclidean norms of flattened action tensors, and $i$ indexes their entries.
When the reference norm is zero, relative error is undefined and we report absolute error.
Both outputs must share a representation and channel units.
Precision changes are evaluated separately from implementation checks; closed-loop rollouts measure their behavioral consequences.

\section{Evaluation}
\label{sec:exp}
\subsection{Experimental setup}
\label{sec:setup}
We measure task success on LIBERO-Object~\cite{libero}, a simulation benchmark with ten language-conditioned manipulation tasks.
The eleven-model evaluation uses 100 episodes per model (Table~\ref{tab:fidelity}); a separate seven-model device study uses 200 episodes per model and its own execution horizons (Table~\ref{tab:tiers}).
Incomplete episodes count as failures in both studies.
Reference rates for $\pi_0$ and GR00T-N1.5/N1.6/N1.7 come from our PyTorch reproductions.
Other reference rates come from published results and provide context rather than matched-checkpoint comparisons.

We report 95\% Wilson confidence intervals for success proportions.
These intervals describe episode-level binomial uncertainty within the evaluated tasks; they exclude variation across training seeds and task distributions.
Success-rate comparisons are descriptive; numerical agreement is checked separately on fixed inputs.

Latency is measured at three boundaries: model execution, server prediction, and the client's complete request--response cycle.
For latency $L$ seconds at the stated measurement boundary and $S$ executed actions per request, the amortized inference time is $L/S$ and the corresponding throughput is $S/L$.
These quantities exclude the time spent executing actions on the robot.
Memory measurements distinguish GPU allocation, process resident set size (RSS), and total system memory use.
On Jetson, the CPU and GPU share memory, and process RSS alone omits some accelerator allocations.

The cross-platform profiles use an RTX 3060 with 12\,GB memory, an Apple M4 with 24\,GB memory, and an Intel Arc integrated GPU through OpenVINO.
The device study additionally evaluates Jetson AGX Orin and the 8\,GB Orin Nano.
Model subsets and timing statistics are specified for each experiment.

\begin{table*}[!t]
\centering
\caption{Eleven-model LIBERO-Object evaluation (100 episodes/model) and separate platform profiles. Success: percent [95\% Wilson interval]; Ref.: published rate or PyTorch reproduction ($^{*}$). Input: image side length (pixels); $S$: executed actions per request. Timing: request P90 (90th percentile)/vision-stage latency (ms). CUDA memory: device allocation/host RSS; Metal and OpenVINO (OV): RSS (MiB). Devices: RTX 3060, Apple M4, and Intel Arc iGPU. --: unmeasured.}
\label{tab:fidelity}
\scriptsize
\setlength{\tabcolsep}{2pt}
\begin{tabular*}{\textwidth}{@{\extracolsep{\fill}}lccrcccrcr@{}}
\toprule
 & & & & \multicolumn{2}{c}{CUDA} & \multicolumn{2}{c}{Metal} & \multicolumn{2}{c}{OV} \\
\cmidrule(lr){5-6}\cmidrule(lr){7-8}\cmidrule(lr){9-10}
Model & Input/$S$ & Success [95\% CI] & Ref. & P90/Vision & GPU/Host & P90/Vision & RSS & P90/Vision & RSS \\
\midrule
SmolVLA & 512/1 & 96 [90.2, 98.4] & 96.0 & 98.9/47.7 & 1322/735 & 330.6/186.3 & 1380 & 391.6/44.8 & 1069 \\
Evo-1 & 448/8 & 96 [90.2, 98.4] & 97.7 & 180.0/56.6 & 1574/822 & 775.7/328.0 & 1620 & 585.7/183.6 & 1200 \\
$\pi_0$ & 224/50 & 82 [73.3, 88.3] & 79.0$^{*}$ & 203.6/40.2 & 5480/833 & 1123.5/189.8 & 5530 & 779.6/83.9 & 5060 \\
$\pi_{0.5}$ & 224/10 & 96 [90.2, 98.4] & 98.2 & 254.1/45.9 & 5938/729 & 1198.2/198.1 & 5987 & 568.1/37.7 & 5490 \\
GR00T-N1.5 & 224/16 & 95 [88.8, 97.8] & 95.0$^{*}$ & 95.2/23.2 & 4830/1436 & 423.2/98.9 & 6841 & 249.1/28.3 & 5550 \\
GR00T-N1.6 & 224/16 & 88 [80.2, 93.0] & 99.0$^{*}$ & 91.3/23.8 & 5918/1441 & 389.9/99.8 & 7144 & 167.3/37.7 & 4480 \\
GR00T-N1.7 & 256/16 & 99 [94.6, 99.8] & 99.0$^{*}$ & 93.8/21.7 & 6288/1417 & 385.0/87.3 & 7442 & 299.8/56.8 & 5890 \\
VLA-JEPA & 224/7 & 100 [96.3, 100] & 99.6 & 64.2/22.3 & 3986/627 & 258.2/87.3 & 4034 & 151.2/55.3 & 3690 \\
VLA-Adapter & 224/8 & 96 [90.2, 98.4] & 99.2 & 82.1/44.1 & 2864/1063 & 332.4/189.1 & 3697 & 239.3/121.5 & 2590 \\
OpenVLA-OFT & 224/8 & 97 [91.5, 99.0] & 98.4 & --/-- & --/-- & 1719.9/194.0 & 15457 & 482.7/125.2 & 14040 \\
BitVLA & 224/8 & 100 [96.3, 100] & 99.6 & 62.0/12.0 & 1312/1072 & --/-- & -- & --/-- & -- \\
\bottomrule
\end{tabular*}
\end{table*}

\subsection{Task success and hardware coverage}
\label{sec:fidelity}
Table~\ref{tab:fidelity} covers all eleven supported models.
Ten achieve success rates within 3.2 percentage points of the corresponding reference rate.
GR00T-N1.6 achieves 88\% against 99\% for its PyTorch reference; the cause of this gap remains unresolved.
Similar success rates do not establish numerical equivalence, particularly when reference protocols differ.
The reference rates for SmolVLA, Evo-1, VLA-JEPA, VLA-Adapter, OpenVLA-OFT, and BitVLA come from their model reports~\cite{smolvla,evo1,vlajepa,vlaadapter,openvlaoft,bitvla}; the $\pi_{0.5}$ rate is the OpenPI reference reported in LeRobot's documentation~\cite{lerobotpi05}.

\paragraph{Additional tasks}
A separate evaluation of BitVLA and GR00T-N1.7 uses 200 episodes per model per suite across all four LIBERO suites (Table~\ref{tab:libero-full}).
Both obtain their lowest rate on Long, which requires longer task sequences.
On SimplerEnv's WidowX tasks~\cite{simplerenv}, GR00T-N1.6 at 252-pixel input achieves 80\% on ``put carrot on plate'' and 70\% on ``put spoon on towel.''

\begin{table}[!t]
\centering
\caption{Four-suite evaluation: success (\%) [95\% Wilson interval], 200 episodes/model/suite. BitVLA published rates~\cite{bitvla} are contextual references.}
\label{tab:libero-full}
\small\setlength{\tabcolsep}{3pt}
\begin{tabular}{lrrr}
\toprule
Suite & BitVLA & Ref. & GR00T-N1.7 \\
\midrule
Spatial & 94.5 [90.4, 96.9] & 97.4 & 96.0 [92.3, 98.0] \\
Object & 100 [98.1, 100] & 99.6 & 98.0 [95.0, 99.2] \\
Goal & 92.5 [88.0, 95.4] & 94.4 & 97.5 [94.3, 98.9] \\
Long & 83.5 [77.7, 88.0] & 87.6 & 89.0 [83.9, 92.6] \\
\bottomrule
\end{tabular}
\end{table}

\paragraph{Backend coverage}
Table~\ref{tab:fidelity} includes CUDA, Metal, and OpenVINO profiles, with coverage varying by model.
CPU checks on an eight-thread i9-14900HX return actions in 2141, 2931, 3341, 4888, 7888, and 8966\,ms for SmolVLA, GR00T-N1.7, N1.6, N1.5, $\pi_0$, and Evo-1, respectively, at 1.4--6.1\,GiB peak RSS.
The tested packed BitVLA configuration uses CUDA.
Separate M4 Metal trials yield 374\,ms and 9/10 successes for SmolVLA, 1164\,ms and 8/10 for $\pi_0$, and 539\,ms and 10/10 for GR00T-N1.7.
On an Intel Arc A380, SYCL reduces request latency relative to its Ryzen 5 host from 1920 to 630\,ms for SmolVLA, 7695 to 1176\,ms for Evo-1, and 2994 to 517\,ms for VLA-Adapter.
Maximum absolute action difference is $2.9\times10^{-3}$ against CPU outputs with magnitude up to 0.99.

\begin{table}[!t]
\centering
\caption{Seven-model device study with platform-specific software revisions. Platform cells give client inference time per executed action $L/S$ (ms), including preprocessing and transport, followed by successes out of 200. $\pi_0$: $S=32$ on RTX/AGX, $S=50$ on Nano. RSS: Nano peak process RSS (MiB). $^{\dagger}$Remote simulator. OOM: out of memory.}
\label{tab:tiers}
\footnotesize
\setlength{\tabcolsep}{2pt}
\begin{tabular}{lrrrrr}
\toprule
Model & $S$ & RTX 3060 & AGX Orin & Orin Nano & RSS \\
\midrule
SmolVLA & 4 & 28.16/181 & 65.41/181 & 141.81/176 & 2031 \\
BitVLA & 8 & 37.85/200 & 101.11/200 & 355.65/200 & 2199 \\
Evo-1 & 8 & 63.60/189 & 131.01/191 & 458.84/195 & 2135 \\
GR00T-N1.5 & 16 & 14.17/192 & 28.78/195 & 84.76/192$^{\dagger}$ & 5975 \\
$\pi_0$ & 32/50 & 9.74/175 & 27.90/171 & 39.10/161$^{\dagger}$ & 6068 \\
GR00T-N1.6 & 16 & 10.29/173 & 26.70/180 & OOM & -- \\
GR00T-N1.7 & 16 & 10.26/196 & 26.84/197 & OOM & -- \\
\bottomrule
\end{tabular}
\end{table}

\subsection{Inference latency}
\label{sec:speedup}
\label{sec:baselines}
Table~\ref{tab:baselines-full} compares seven models on an RTX 3090 against eager PyTorch and \texttt{torch.compile} in default mode.
Eager execution dispatches operations as the model runs; compilation optimizes groups of operations before execution.
The comparison holds checkpoint, precision, and inputs fixed and averages 200 timed calls after warmup, excluding transport and image serialization.
Relative to eager execution, the measured speedup is $1.02$--$6.21\times$.
Compilation closes much of this gap.
Against the faster of the two PyTorch configurations, \texttt{vla.cpp} achieves similar latency for four models and speedups of $1.18\times$, $1.26\times$, and $1.84\times$ for SmolVLA, GR00T-N1.5, and BitVLA, respectively.

For GR00T-N1.7, compilation allows eager fallback via \texttt{suppress\_errors}~\cite{pytorchcompile}.
The timing includes any fallback taken by that configuration.

\begin{table}[!t]
\centering
\caption{Server-side latency (ms) on RTX 3090: mean of 200 calls after warmup. Compile: PyTorch default mode. $^{\ddagger}$Eager fallback enabled. Speedup: faster PyTorch latency divided by ours.}
\label{tab:baselines-full}
\footnotesize
\setlength{\tabcolsep}{3.5pt}
\begin{tabular}{lrrrr}
\toprule
Model & Ours & Eager & Compile & Speedup \\
\midrule
SmolVLA & 55.8 & 174.1 & 65.9 & 1.18$\times$ \\
$\pi_0$ & 97.9 & 99.4 & 99.7 & 1.02$\times$ \\
Evo-1 & 142.3 & 161.6 & 142.5 & 1.00$\times$ \\
GR00T-N1.5 & 75.4 & 100.5 & 94.7 & 1.26$\times$ \\
GR00T-N1.6 & 61.2 & 76.0 & 61.1 & 1.00$\times$ \\
GR00T-N1.7 & 59.9 & 66.1 & 60.1$^{\ddagger}$ & 1.00$\times$ \\
BitVLA & 51.5 & 319.7 & 94.6 & 1.84$\times$ \\
\bottomrule
\end{tabular}
\end{table}

A separate RTX 5070 test (median of 100 predictions, transport excluded) gives 74.11\,ms for SmolVLA versus 175.99\,ms with eager PyTorch and 76.26\,ms with graph-captured PyTorch.
GR00T-N1.7 gives 101.38\,ms versus 101.67\,ms with eager PyTorch; no captured result was obtained in that configuration.

A matched GR00T-N1.7 comparison on an RTX PRO Blackwell 1000 favors TensorRT: median latency is $70.5$\,ms against $130.2$\,ms for \texttt{vla.cpp}, a $1.85\times$ speedup.
Success counts are 98/100 and 96/100, respectively, with checkpoint, input configuration, action horizon, and precision held fixed.
TensorRT provides the faster deployment for this configuration, while \texttt{vla.cpp} provides a shared implementation across processor backends.

\subsection{Memory footprint and ternary acceleration}
\label{sec:memory}
Memory capacity limits which policies can run onboard.
Five of the seven models in Table~\ref{tab:tiers} load on the 8\,GB Orin Nano.
GR00T-N1.6/N1.7 exceed the available memory in this configuration, and GR00T-N1.5 and $\pi_0$ require the simulator to run on another machine.

BitVLA completes 200/200 LIBERO-Object episodes on the Nano, with 2199\,MiB peak process RSS.
Peak system memory use is 6504\,MiB, an increase of 4533\,MiB over the initial baseline.
These system measurements include the surrounding workload and cannot be attributed entirely to model execution.
On RTX 3060 the measured device allocation is 1312\,MiB ($1.28$\,GiB).
Process RSS therefore understates the Nano deployment's total memory use.

Packing ternary weights reduces the model file from $5.6$ to $1.34$\,GiB and transient host allocation during loading from $11.1$ to $1.15$\,GiB.
This storage change leaves latency unchanged when the arithmetic implementation is held fixed.
Replacing the \texttt{dp4a} path with the IMMA implementation reduces client inference time per executed action from $172.8$ to $37.85$\,ms on RTX 3060 and from $406.6$ to $101.11$\,ms on AGX Orin: $4.57\times$ and $4.02\times$, respectively, at the same replay setting.
Both comparisons use $S=8$ and include client processing and transport.
Task success remains 100\%.
Thus compact storage reduces the loading footprint, while tensor-core execution reduces inference time.

\paragraph{Other weight precisions}
Table~\ref{tab:quant} evaluates GGUF Q8\_0 and Q4\_0 quantization for GR00T-N1.7, including variants that also quantize the vision tower.
The most compact file is 4556.0\,MB versus 6292.4\,MB for BF16, with a $1.144\times$ fixed-input speedup.
Plain Q4\_0 achieves 190/200 successes versus 196/200 for BF16: a $-3.0$ percentage-point difference with an approximate 95\% interval of $[-7.1,+0.8]$ under independent sampling.
Q4\_0 with vision quantization reaches 195/200 despite similar relative RMS error.
Fixed-input error alone does not predict these task-success rates.

\begin{table}[!t]
\centering
\caption{GR00T-N1.7 quantization, RTX 3060. Success: LIBERO-Object, 200 episodes/configuration, $S=16$. Speedup and relative RMS error $e_{\mathrm{rel}}$ use fixed inputs; error compares 5280 action values with BF16. +V also quantizes vision.}
\label{tab:quant}
\footnotesize\setlength{\tabcolsep}{3pt}
\begin{tabular}{lrrrr}
\toprule
Weights & File (MB) & Speedup ($\times$) & $e_{\mathrm{rel}}$ & Success \\
\midrule
BF16 & 6292.4 & 1.000 & 0 & 196/200 \\
Q8\_0 & 5537.4 & 1.077 & 0.0043 & 197/200 \\
Q4\_0 & 5134.8 & 1.081 & 0.0480 & 190/200 \\
Q8\_0+V & 5160.0 & 1.131 & 0.0050 & 195/200 \\
Q4\_0+V & 4556.0 & 1.144 & 0.0490 & 195/200 \\
\bottomrule
\end{tabular}
\end{table}

\subsection{Solver steps and execution costs}
\label{sec:roofline}
\begin{table}[!t]
\centering
\caption{GR00T-N1.7 solver sweep, RTX 3060. Success: runtime/PyTorch (\%), 100 episodes per stack and setting. Times: vision/backbone-plus-head (ms). Relative RMS error $e_{\mathrm{rel}}$ compares 5280 fixed-input action values with the runtime's $T=4$ output.}
\label{tab:solver-sweep}
\footnotesize\setlength{\tabcolsep}{5pt}
\begin{tabular}{rrrr}
\toprule
$T$ & Success & Times (ms) & $e_{\mathrm{rel}}$ \\
\midrule
1 & 99/96 & 18.49/19.60 & 0.2290 \\
2 & 95/98 & 18.83/25.79 & 0.0684 \\
4 & 98/94 & 18.68/37.21 & 0 \\
8 & 98/95 & 18.68/60.54 & 0.0321 \\
16 & 95/97 & 18.86/107.50 & 0.0483 \\
\bottomrule
\end{tabular}
\end{table}

\begin{figure*}[!t]
\centering
\includegraphics[width=0.85\textwidth]{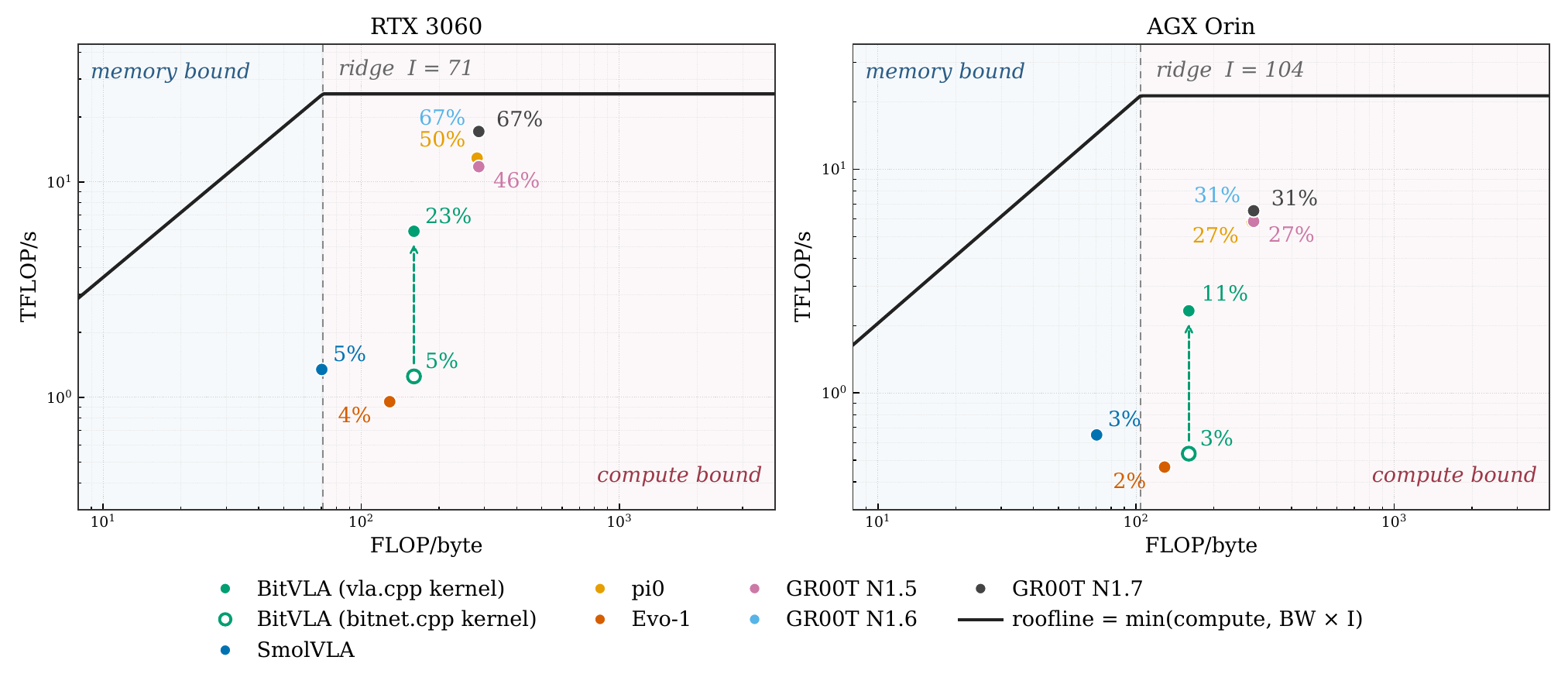}
\caption{Estimated arithmetic intensity and throughput. Compute ceilings are measured GEMM rates of 25.6 and 21.3\,TFLOP/s; bandwidth ceilings are 360 and 204.8\,GB/s for RTX 3060 and AGX Orin. Point labels give achieved throughput as a fraction of the ceiling. Placement depends on FLOP and byte accounting.}
\label{fig:roofline}
\end{figure*}

To examine the fixed and repeated costs in Eq.~\eqref{eq:cost}, we vary GR00T-N1.7's solver steps over $T\in\{1,2,4,8,16\}$ on RTX 3060 (Table~\ref{tab:solver-sweep}).
Vision time remains between $18.5$ and $18.9$\,ms, while backbone-plus-head time increases from $19.6$ to $107.5$\,ms.
A linear fit to the latter gives $13.9+5.8T$\,ms.
Including vision, the fit corresponds to approximately $32.6$\,ms of fixed work and $5.8$\,ms per solver step within the tested range.
Client-side fits are approximately $54+6.2T$\,ms for \texttt{vla.cpp} and $67+7.3T$\,ms for PyTorch.
Success ranges from 95/100 to 99/100 for the runtime and from 94/100 to 98/100 for PyTorch across the sweep.
Relative RMS error against the $T=4$ output is 0.229 at $T=1$ and 0.032--0.048 at $T\in\{8,16\}$; this comparison measures sensitivity to solver steps, not agreement with PyTorch.
At $T=1$, the maximum absolute action difference from the $T=4$ output is 0.93 despite 99/100 successful episodes.

The balance between fixed and repeated work varies by model.
On RTX 5070, vision and backbone respectively account for 38\% and 14\% of SmolVLA execution time ($T=10$), 31\% and 21\% for GR00T-N1.6 ($T=4$), and 21\% and 54\% for $\pi_0$ ($T=10$).
The combined shares are 52\%, 52\%, and 75\%; for the direct-head BitVLA, the share is approximately 99.5\%.
The action expert thus takes nearly half the time for SmolVLA and GR00T-N1.6 even with cached observation features.
GR00T-N1.7's separate RTX 3060 sweep attributes 58\% to vision plus backbone and 42\% to the expert at $T=4$.

Roofline analysis~\cite{roofline,vlaperf} relates these costs to \emph{arithmetic intensity}, the number of operations per byte transferred.
For intensity $I$ in FLOPs/byte, the roofline model bounds throughput by $\min(P_{\max},BI)$, where $B$ is memory bandwidth in bytes/s and $P_{\max}$ is peak arithmetic throughput in FLOP/s.
Dense backbone operations reuse weights across many tokens, whereas a small action expert may have lower intensity.
Figure~\ref{fig:roofline} places the models relative to measured matrix-multiplication (GEMM) ceilings and nominal bandwidth.
The resulting ridge points are 71 and 104\,FLOPs/byte for RTX 3060 and AGX Orin; bandwidth estimates of 300 and 150\,GB/s move them to 85 and 142.
Prefix intensity estimates of 256--530\,FLOPs/byte exceed these ridges, whereas the expert estimate is approximately 50.
These estimates depend on FLOP and memory-traffic accounting; proximity to a ceiling alone does not establish resource saturation.

\begin{figure*}[!t]
\centering
\includegraphics[width=0.85\textwidth]{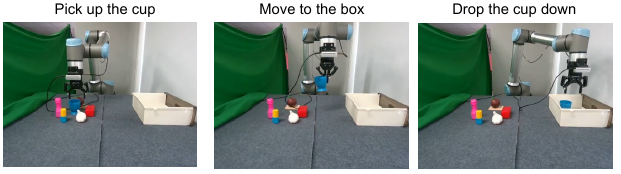}
\caption{UR10e cup pick-and-place task: grasp the cup, carry it to the basket, and release it. Across the 80 evaluated rollouts, all failures occur at release.}
\label{fig:ur10e}
\end{figure*}

\subsection{Sensitivity to positional-index precision}
\label{sec:precision}
SmolVLA illustrates how a small implementation difference can change policy behavior.
Its vision encoder maps image-patch coordinates to discrete indices, which select learned positional embeddings.
Computing the coordinates at a different precision can move a value across an index boundary and select another embedding.
Matching the half-precision reference reduces the position-embedding relative error from $0.32$ to $0.003$.
On LIBERO-Object task 0, placing a soup container in a basket, success improves from 2/10 to 9/10 episodes.

The maximum absolute action difference is approximately $1.97$ after denormalization, concentrated on the gripper channel.
This magnitude is large enough to reverse the gripper command.
Matching the index precision also reduces this maximum difference below $10^{-2}$.
The values are simulator command units, not physical displacements.
A separate SmolVLA solver diagnostic finds the maximum absolute action difference rising from 0.26 at one Euler step to 1.05 at eight; it uses a different configuration from the 1.97 maximum.
The example shows why numerical validation must include discrete preprocessing decisions as well as floating-point network operations.

For comparison, GR00T-N1.6 uses interpolated position embeddings and achieves 10/10 successes in both tested precision configurations, with an action difference of roughly $0.003$.
The policies differ in weights and architecture as well as positional operations.
A Metal precision check reduces fixed-input action discrepancy from $2.0\times10^{-3}$ to $8.3\times10^{-5}$ by using 32-bit vision and language towers, adding 33\,ms and 672\,MiB.
These cases motivate checking the complete numerical path on each backend.

\subsection{Physical deployment and execution horizon}
\label{sec:robot}
We evaluate the runtime on a UR10e manipulator and an ALOHA robot using synchronous chunked control.

\paragraph{UR10e deployment}
A UR10e runs SmolVLA fine-tuned on 80 demonstrations of the cup pick-and-place task in Fig.~\ref{fig:ur10e}.
The client executes $S=25$ actions at 20\,Hz, then waits for the next chunk.
Across four backends, round-trip latency ranges from $111.4$ to $3652.9$\,ms, approximately $33\times$ (Table~\ref{tab:robot}).
The effective rate, measured as executed actions per wall-clock second, falls from $17.7$ to $5.0$\,Hz.
In a blocking loop, each cycle contains both the request time and the action-execution time, giving
\begin{equation}
 f_{\mathrm{exec}}\approx\frac{S}{L_{\mathrm{rt}}+S\Delta t},
 \label{eq:control}
\end{equation}
where $L_{\mathrm{rt}}$ is round-trip latency and $\Delta t$ is the interval between actions; other overhead is neglected.
This rate includes robot execution, unlike inference throughput $S/L_{\mathrm{rt}}$.

\begin{table}[!t]
\centering
\caption{UR10e deployment with SmolVLA: $S=25$ and 20 episodes per backend. Success: count/20 [95\% Wilson interval in \%]. $f_{\mathrm{exec}}$: executed actions per wall-clock second. Metal and Ryzen include an approximately 161\,ms network component. Metal uses 32-bit floating-point vision and language computations.}
\label{tab:robot}
\footnotesize
\setlength{\tabcolsep}{2.5pt}
\begin{tabular}{lrrr}
\toprule
Backend & Round trip (ms) & Success [95\% CI] & $f_{\mathrm{exec}}$ (Hz) \\
\midrule
CUDA, RTX 5070 laptop & 111.4 & 15/20 [53.1, 88.8] & 17.7 \\
Metal, Mac Mini M4 & 582.0 & 12/20 [38.7, 78.1] & 12.8 \\
CPU, i9-14900HX & 2164.4 & 12/20 [38.7, 78.1] & 7.1 \\
CPU, Ryzen 7 6850HS & 3652.9 & 14/20 [48.1, 85.5] & 5.0 \\
\bottomrule
\end{tabular}
\end{table}

Success ranges from 12/20 to 15/20 across backends.
All 80 rollouts grasp and carry the cup to the basket; the 27 failures occur at release.
This common failure mode reduces the sensitivity of binary success to latency effects.
Mean steps to grasp are $335\pm26$, $374\pm44$, $329\pm41$, and $348\pm26$ for CUDA, Metal, i9, and Ryzen, respectively (mean $\pm$ standard deviation).
The fraction of wall time blocked on prediction increases from 8\% to 73\% between the fastest and slowest settings.
The experiment demonstrates cross-backend operation, with differences in machine, network path, and, for Metal, numerical precision.
In a separate same-machine engine comparison, a $1.56\times$ latency increase coincides with 111 fewer steps to first grasp (Welch $t=-8.7$, 27 degrees of freedom).
The delay-injection study below holds hardware and precision fixed.

An embedded ALOHA deployment runs GR00T-N1.6 on AGX Orin at approximately $470$\,ms per predicted chunk, with $H=8$ and $S=7$.
The first task places all blocks on the table into a white box; the second places trash into a box and then a banana onto a white dish.
They achieve 18/20 and 17/20 successes, respectively, with mean completion times of approximately 49 and 28\,s over successful trials and a 1.3\,s replan period.

\begin{figure*}[!t]
\centering
\begin{minipage}{0.52\textwidth}
\includegraphics[width=\linewidth]{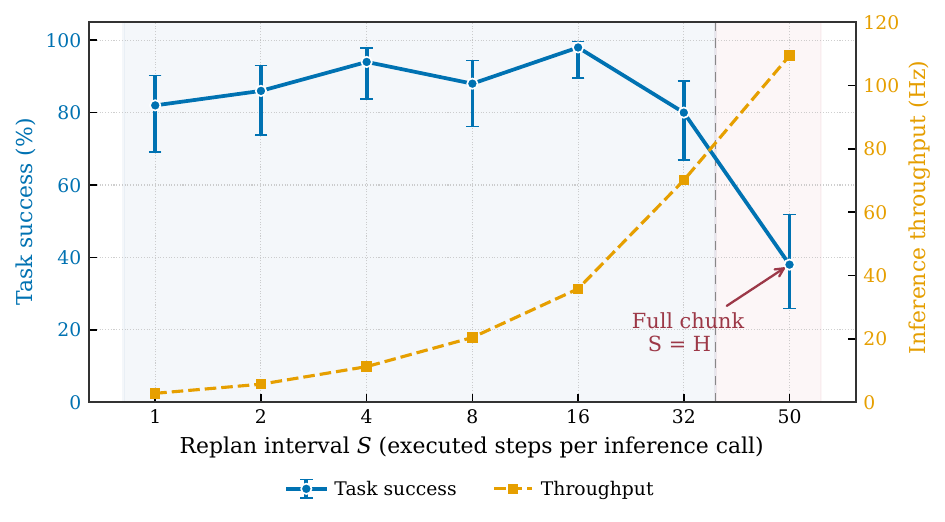}
\end{minipage}\hfill
\begin{minipage}{0.46\textwidth}
\centering\footnotesize\setlength{\tabcolsep}{3pt}
\begin{tabular}{rrrrr}
\toprule
$S$ & Median & P95 & $S/L$ & Success \\
 & (ms) & (ms) & (Hz) & (\%) \\
\midrule
1 & 358.4 & 370.5 & 2.8 & 82 \\
2 & 352.7 & 363.4 & 5.7 & 86 \\
4 & 356.6 & 383.2 & 11.2 & 94 \\
8 & 391.7 & 429.9 & 20.4 & 88 \\
16 & 447.0 & 463.3 & 35.8 & 98 \\
32 & 456.7 & 466.8 & 70.1 & 80 \\
50 & 457.0 & 467.9 & 109.4 & 38 \\
\bottomrule
\end{tabular}
\end{minipage}
\caption{SmolVLA execution-horizon sweep: LIBERO-Object on Orin Nano, $H=50$, 50 episodes/setting. Error bars are 95\% Wilson intervals. $L$ is the reported inference latency, excluding transport; throughput $S/L$ also excludes action execution. P95: 95th percentile. At $S=H$, the policy receives new observations between chunks but not within a chunk.}
\label{fig:replan}
\end{figure*}
\paragraph{Controlled latency injection}
To isolate inference delay from hardware and precision changes, we insert an artificial delay after GR00T-N1.7 inference on an RTX 4070 and before delivering each chunk to ALOHA.
Table~\ref{tab:delay} reports ten episodes per condition.
Success remains 9/10 or 10/10 over delays from 0 to 2000\,ms; average executed steps do not increase monotonically.
The ten-episode samples provide limited sensitivity to changes near 100\% success.
Executed steps also exclude time spent waiting, so similar step counts do not imply similar completion times.

\begin{table}[!t]
\centering
\caption{Controlled delay injection on ALOHA with GR00T-N1.7 (RTX 4070). Delay is added after inference. Each condition uses ten episodes. Steps: mean executed actions per episode, excluding waiting time.}
\label{tab:delay}
\footnotesize
\begin{tabular}{rrrr}
\toprule
Delay (ms) & Steps & Success & 95\% CI (\%) \\
\midrule
0 & 97.6 & 10/10 & [72.2, 100] \\
150 & 93.2 & 10/10 & [72.2, 100] \\
330 & 90.8 & 10/10 & [72.2, 100] \\
530 & 96.8 & 9/10 & [59.6, 98.2] \\
1000 & 99.6 & 9/10 & [59.6, 98.2] \\
2000 & 92.4 & 10/10 & [72.2, 100] \\
\bottomrule
\end{tabular}
\end{table}

\paragraph{Execution horizon in simulation}

The execution horizon controls how long the robot acts before receiving new visual feedback.
We vary $S$ for SmolVLA on LIBERO-Object with $H=50$, holding the checkpoint, seed, and observation pipeline fixed and evaluating 50 episodes per setting on Orin Nano (Fig.~\ref{fig:replan}).
Median inference latency ranges from 353 to 457\,ms.
Success varies non-monotonically for $S\leq32$, reaching 49/50 at $S=16$, and falls to 19/50 at $S=50$.
At the latter setting, the robot executes the complete predicted chunk before observing again, although it still replans between chunks.

An exploratory two-sided Fisher exact test gives $p=0.0078$ for $S=16$ versus $S=32$, assuming independent episodes and before adjustment for multiple comparisons.
The decline at $S=50$ supports refreshing observations within the predicted horizon for this policy and task suite.
Increasing $S$ amortizes inference over more actions, but also delays feedback.
Runtime comparisons should therefore specify both $S$ and solver steps $T$, since they affect different parts of the computation and control loop.

\section{Limitations and Conclusion}
\label{sec:conclusion}
Evaluation coverage varies across models and backends, and published reference rates may use different checkpoints or protocols.
GR00T-N1.6's LIBERO-Object gap remains unresolved.
The largest tested models exceed the Nano's available memory, the ternary tensor-core path is CUDA-specific, and CPU inference is too slow for many control settings.
The robot studies use small samples in quasi-static scenes; dynamic manipulation requires further evaluation.

\texttt{vla.cpp} provides a common execution system for eleven VLA models across heterogeneous processors.
It preserves architecture-specific computation while sharing model loading, tensor execution, and serving.
The experiments show that compact storage, efficient kernels, and reference-matched computation address distinct requirements of robot deployment.
Separating prediction cost from the execution horizon clarifies the tradeoffs when selecting a policy, hardware, and control configuration.

\IEEEtriggeratref{16}
\AddToHook{env/thebibliography/begin}{\interlinepenalty=10000}
\bibliographystyle{IEEEtran}
\bibliography{references}
\end{document}